\documentclass[a4paper]{easychair}
\usepackage[english,american]{babel}
\usepackage{ctable}
\usepackage{xspace}
\usepackage{hyperref}

\newcommand{\comment}[1]{}
\newcommand{\essence}{E{\small SSENCE}\xspace}
\newcommand{\buf}{\mathit{Buf}}
\newcommand{\bout}{\mathit{Bout}}

\title{Solving the Wastewater Treatment Plant Problem with SMT}
\titlerunning{Solving the Wastewater Treatment Plant Problem with SMT}

\author{Miquel Bofill, V\'{\i}ctor Mu\~noz, and Javier Murillo\\
University of Girona\\
Girona, Catalonia, Spain\\
\url{miquel.bofill@udg.edu}, \url{{victor.munoz,javier.murillo}@newronia.com}
}
\authorrunning{Bofill, Mu\~noz, and Murillo}

\begin{document}
\maketitle

\begin{abstract}
  In this paper we introduce the Wastewater Treatment Plant Problem, a
  real-world scheduling problem, and compare the performance of
  several tools on it. We show that, for a naive modeling,
  state-of-the-art SMT solvers outperform other tools ranging from
  mathematical programming to constraint programming. We use both real
  and randomly generated benchmarks.

  From this and similar results, we claim for the convenience of
  developing compiler front-ends being able to translate from
  constraint programming languages to the SMT-LIB standard language.
\end{abstract}

\section{Introduction}

Most state-of-the-art satisfiability checkers for propositional logic,
also known as SAT solvers, are based on variations of the
Davis-Putnam-Logemann-Loveland (DPLL)
procedure~\cite{DavisPutnam1960,DavisLogemannLoveland1962}.  During
the last ten years, SAT solvers have spectacularly progressed in
performance thanks to better implementation techniques and conceptual
enhancements, such as non-chronological backtracking and 
conflict-driven lemma learning, which in many instances of real
problems are able to reduce the size of the search space
significantly.  Thanks to those advances, nowadays best SAT solvers
can tackle problems with hundreds of thousands of variables and
millions of clauses.

These techniques have been adapted for more expressive (yet decidable)
logics. For instance, in hardware and software verification
applications, decision procedures for checking the satisfiability of
quantifier-free first-order formulas with respect to background
theories (such as the integer or real numbers, arrays, bit vectors,
etc) have been developed.  This is known as the SAT Modulo
  Theories (SMT) problem for a theory $T$: given a formula $F$,
determine whether there is a model of $T\cup\{F\}$.  Hence, an SMT
instance is a generalization of a Boolean SAT instance in which some
propositional variables have been replaced by predicates from the
underlying theories, and can contain clauses like, e.g., $p\lor q\lor
x+2\leq y\lor x=y+z$, providing a much richer modeling language than
is possible with SAT formulas.

Although an SMT instance can be solved by encoding it into an
equisatisfiable SAT instance and feeding it to a SAT solver, currently
most successful SMT solvers are based on the integration of a SAT
solver and a $T$-solver, that is, a decision procedure for the given
theory $T$. In this so-called lazy approach, while the SAT solver is
in charge of the Boolean component of reasoning, the $T$-solver deals
with sets of atomic constraints in $T$. The main idea is that the
$T$-solver analyzes the partial model that the SAT solver is building,
and warns it about conflicts with theory $T$
($T$-inconsistency). See~\cite{Sebastiani2007} for a survey on this
approach.

Leveraging the advances made in SAT solvers in the last decade, SMT
solvers have proved to be competitive with classical decision methods
in many areas. See, e.g., \cite{NieuwenhuisOliveras2006SAT} for
an application of an SMT solver on an optimization problem, being
competitive with the best weighted CSP solver with its best heuristic
on that problem.
In the spirit of bringing SMT technology to other communities, in this
paper we show how a non-trivial scheduling problem can be efficiently
solved by using state-of-the-art SMT solvers, and compare their
performance to other approaches.  The problem we deal with is a
real-world cumulative scheduling problem, namely, the Wastewater
  Treatment Plant Problem (WWTPP). This problem turns out to be a
generalization of the preemptive CuSP.

The rest of the paper is organized as follows. In Section~\ref{cusp}
we give some preliminaries on cumulative scheduling. In
Section~\ref{wwtpp} we define the WWTPP and, in Section~\ref{modeling}
we encode it by translation into SAT modulo Linear Integer Arithmetic
and also into Integer Programming (IP). In Section~\ref{benchmarking}
we give computational results comparing the performance of different
tools on the problem, both for real and random instances, which
suggest that state-of-the-art SMT solvers are competitive with best IP
tools, and even better on difficult instances of this problem.  For
the sake of completeness, in Section~\ref{cp} the problem is encoded
into many Constraint Programming (CP) dialects, preserving as much as
possible the encoding given for SMT. We give computational results
showing that SMT tools have far much better performance than CP tools
on this problem. After this, in Section~\ref{cum}, a different
encoding of the problem is discussed for the CP tools supporting the
global {\tt cumulative} constraint.  Overall conclusions and ideas for
future work are given in Section~\ref{conclusion}.

\section{Cumulative Scheduling}\label{cusp}

Scheduling problems are prototypical NP-complete real-world
problems. Roughly speaking, scheduling is the process of assigning
scarce resources to tasks over time. It is not surprising, hence, that
those problems have been addressed in a wide variety of research
areas, and that they have been divided into different subclasses. Many
real-world scheduling problems fall into the cumulative scheduling
class, where each resource may be shared between a bounded number of
tasks~\cite{AggounBeldiceanu93a}, in contrast with the disjunctive
scheduling class, where each resource can execute at most one task at
a time. Another frequent differentiation is between preemptive and
non-preemptive scheduling, depending on whether a task can be
interrupted (i.e., subdivided) or not.

A very general problem in the cumulative scheduling class is the
resource-constrained project sche\-duling problem (RCPSP). In its
non-preemptive version, we have (1) a set of resources of given
capacities, (2) a set of tasks of given durations, (3) an acyclic
graph of precedence constraints between the tasks, and (4) for each
task and each resource the amount of the resource required by the task
over its execution. The objective is to find a start time assignment
to the tasks that satisfies the precedences and resource capacity
constraints, and minimizes the makespan (i.e., the time at which all
tasks are completed). As a generalization of the job-shop scheduling
problem, the decision variant of the RCPSP (where what is required is
to find an assignment whose makespan does not exceed a given deadline)
is NP-complete in the strong sense and, hence, the RCPSP is NP-hard in
the strong sense~\cite{GareyJohnson79,Blazewiczetal83}.
See~\cite{HerroelenDemeulemeester02,BruckerKnust06} for
surveys.

Due to the generality of the RCPSP, many problems in the cumulative
scheduling class are particular cases of it. This is the case for the
so-called cumulative scheduling problem (CuSP), where we have (1) a
single resource of given capacity and (2) a set of tasks, each one
with a given duration, release time, deadline and resource capacity
requirement, and we are asked to find an schedule satisfying all
timing constraints and the resource capacity
constraint~\cite{Baptisteetal99}. As an extension of the parallel
machine problem~\cite{CarlierThesis}, this problem is also NP-hard in
the strong sense.

Most of the work on the RCPSP and the CuSP has been devoted to the
non-preemptive case (note that the introduction of preemption
increases the number of possible solutions). 
Some relaxations of the CuSP involving elastic tasks (where, roughly,
each task is only given a global amount of processing, and its
duration and consumption is not fixed) have been considered
in~\cite{CaseauLaburthe96,Baptisteetal99}, but little work on the
preemptive case can be found in the literature. See, e.g.,
\cite{KaplanThesis,DemeulemeesterHerroelen96} for the preemptive RCPSP
and~\cite{CarlierNeron00} for the preemptive CuSP.

In this paper we deal with the Wastewater Treatment Plant Problem
(WWTPP), a real-world cumulative scheduling problem which is a
generalization of the preemptive CuSP.

\section{The Wastewater Treatment Plant Problem (WWTPP)}\label{wwtpp}

The treatment of the wastewater discharged by industries into the
rivers is vital for environmental quality. For this purpose, the
wastewater is treated in wastewater treatment plants (WWTP). A WWTP
receives the polluted wastewater discharges coming from the city and
different industries.  Nowadays the most common wastewater treatment
is the activated sludge process. The system consists in an aeration
tank in which the microorganisms responsible for treatment
(i.e. removal of carbon, nitrogen and phosphorous) are kept in
suspension and aerated followed by a liquid-solids separation, usually
called secondary settler. Finally a recycle system is responsible for
returning a fraction of solids removed from the liquid-solids
separation unit back to the reactor, whereas the other fraction is
wasted from the system~\cite{WWTP}.

The treatment capacity of the plant is limited, and therefore all
pollutants arriving at the WWTP should be under certain limits;
otherwise, the wastewater could not be fully treated and the river
would be polluted. The hydraulic and contaminants capacity constraints
that are defined according to its expected use (industries and cities
in the surroundings that generate the waste). Currently, there exist
regulations intended to achieve this goal by assigning a fixed amount
of authorized discharges to each industry.  However, they are not
sufficient to guarantee the proper treatment of the wastewater. The
problem is that, although these regulations enforce industries to
respect the WWTP capacity thresholds, they do not take into account
that simultaneous discharges by different industries may exceed the
WWTP's thresholds. In such a case, no industry would be breaking the
rules, but the effect would be to exceed the WWTP capacity.

The scheduling problem faced in this domain is to distribute the
industrial discharges over time so that all the water entering the
WWTP can be treated.  If the discharges are done without any
coordination, the amount of water arriving at the WWTP can exceed its
incoming water flow threshold, and cause the overflow to go directly
to the river without being treated. Moreover, if the contamination
level of the water is too high, the microorganisms used in the
cleaning process die, and the process has to stop until they are
regenerated. Thus, in order to prevent such dangerous situations, the
industrial discharges should be temporally distributed so that all of
them can be fully treated.

In this paper we address this problem, but taking into account only
the water flow (i.e., without taking into account the contaminants).
Therefore we assume having a single resource of given capacity (the
WWTP) and a set of tasks, each one with a given duration, release
time, deadline and resource capacity requirement (the discharges).

Note that we are asked to find a start time of each discharge between
its release time and its deadline, such that, at any time, the sum of
resource requirements of the discharges scheduled at that time does
not exceed the WWTP capacity. One must assume, however, that there is
some precedence relation (presumably, a chain) between the tasks of
each single industry. Hence, this problem is a generalization of the
CuSP, and a particular case of the RCPSP.

We will focus on the decision variant of the problem, i.e., in finding
a feasible solution not exceeding an overall deadline, instead of in
minimizing the makespan. This is because, in the real problem we
address, it is sufficient that all discharges are rescheduled within
the same day for which they were originally scheduled (and, in fact,
the minimization of the makespan may not be good for the WWTP, as it
is preferably, for the microorganisms' functioning, that the
discharges are homogeneously distributed along the time).  Then we
have that the problem is NP-complete, as it is in NP (as a particular
case of the decision variant of the RCPSP) and it is NP-hard (as a
generalization of the decision variant of the CuSP).
Let us mention that a relaxed version of the WWTPP, where there is no
deadline and conflicts are solved by a recurrent auction mechanism,
has been addressed in~\cite{Munozetal2007}.

Since the delays introduced in the discharges (in order to find a
feasible schedule) should not stop or delay the production processes
of the industries, the idea is to keep those discharges temporarily in
a retention tank in the industry itself, and to throw them to the
river later on.  This approach, however, brings us to a preemptive
framework, since one can reasonably assume that a discharge coming
from a tank can be interrupted (i.e., the tank can be emptied at
disjoint intervals). Hence, that problem has the preemptive CuSP as a
particular case (where, in the CuSP, each task corresponds to a
discharge from a different industry, having a retention tank of
sufficient capacity to hold the discharge and an output rate
equivalent to the one of the discharge), and is NP-hard as well.
Therefore, we define the Wastewater Treatment Plant Problem as
follows.

\subsection{Problem statement}

An instance of the Wastewater Treatment Plant Problem (WWTPP) is
given by (1) a single resource of given capacity, (2) a set of tasks,
each one with a given duration, release time and resource capacity
requirement, (3) a chain-like precedence relation between the tasks,
(4) for any such chain of tasks, a buffer (or retention tank) of given
capacity and output rate (we assume that the input rate is flexible)
and (5) an overall deadline (greater than all release times).

The question is to find an schedule where (1) each
task is either scheduled at its release time (and does not exceed the
deadline), or else it is redirected to its corresponding buffer with a
volume equal to its resource capacity requirement multiplied
by its duration, (2) the capacity of each buffer is not exceeded at
any time, (3) each buffer is emptied, preemptively, at its
corresponding rate, (4) each buffer is empty at the deadline, and (5)
at any time, the sum of required capacities of the tasks scheduled at
that time, together with the required capacities of the emptying of
the buffers at that time, does not exceed the capacity of the single
resource.

Notice that nothing prevents a buffer from being emptied and filled at
the same time, and also from being emptied at the same time at which
one of the tasks is scheduled. 

\section{Modeling the WWTPP}\label{modeling}

In this section we give an encoding of a WWTPP instance into a SAT
modulo unquantified Linear Integer Arithmetic (LIA) instance. As we
will see, SAT modulo LIA nicely captures all constraints.  Afterwards
we translate this encoding into an Integer Programming problem, with
the aim of comparing the performance of state-of-the-art solvers on
both approaches.

\subsection{SMT modeling}\label{SMTmodeling}

A WWTPP instance can be easily codified as a SAT modulo unquantified
Linear Integer Arithmetic instance as follows.

\subsubsection{Constants}
We have the following non-negative integer constants:
\begin{itemize}
\item $\mathit{PlantCapacity}$ denotes the capacity of the wastewater
  treatment plant at each time period.
\item Given a set of $k$ industries, $\mathit{TankCapacity_i}$ and
  $\mathit{TankFlow_i}$ denote respectively the capacity and the
  emptying rate of the buffer associated to industry~$i$, $\forall
  i\in 1\dots k$.
\item Given a set of discharges from $k$ industries to be scheduled
  within $m$ time periods, $d_{ij}$ denotes the scheduled flow of
  discharge for industry $i$ during time period $j$, $\forall i\in
  1\dots k, j\in 1\dots m$.
\end{itemize}

\subsubsection{Variables}

Given a set of discharges from $k$ industries to be scheduled within
$m$ time periods, we have the following integer variables $\forall
i\in 1\dots k, j\in 1\dots m$:
\begin{itemize} 
\item For every $d_{ij}>0$, $c_{ij}$ denotes the actual ``capacity
  requirement'' of industry $i$ during time period $j$, corresponding
  to a scheduled discharge. That is, for every $d_{ij}>0$, either
  $c_{ij}=d_{ij}$, or $c_{ij}=0$ and the discharge is redirected to
  that industry's buffer.
\item $\bout_{ij}$ denotes the flow discharged from buffer (of
  industry) $i$ during time period $j$.
\item $\buf_{ij}$ denotes the flow stored in buffer $i$ at the end of
  time period $j$.
\end{itemize}

\subsubsection{Constraints}

\begin{equation}
\forall j\in 1\dots m:\ \sum_{i=1}^k c_{ij}+\bout_{ij}\leq\mathit{PlantCapacity}\label{1}
\end{equation}

\begin{align}
\forall i\in 1\dots k:\ &\buf_{i1}=d_{i1}-c_{i1}\label{2}\\
\forall i\in 1\dots k, j\in 2\dots
m:\ &\buf_{ij}=\buf_{ij-1}-\bout_{ij}+d_{ij}-c_{ij}\label{3}\\
\forall i\in 1\dots k, j\in 2\dots
m-1:\ &\buf_{ij}\leq\mathit{TankCapacity_i}\label{4}\\
 \forall i\in 1\dots k:\ &\buf_{im}=0\label{5}
\end{align}

In constraints~\ref{2} and~\ref{3}, the difference $d_{ij}-c_{ij}$
is replaced by $0$ if $d_{ij}=0$ (recall that variables $c_{ij}$ have
been defined only for corresponding constants $d_{ij}>0$).

\begin{align}
\forall i\in 1\dots k:\ &\bout_{i1}=0\label{6}\\
\forall i\in 1\dots k, j\in 2\dots
m:\ &\bout_{ij}=0\label{7}\\
&\lor(\bout_{ij}=\mathit{TankFlow_i}
\land \buf_{ij-1}\geq\mathit{TankFlow_i})\label{8}\\
&\lor (\bout_{ij}=\buf_{ij-1} \land \buf_{ij-1}\leq
\mathit{TankFlow_i})\label{9}
\end{align}

For every discharge from an industry $i$, spanning from time period
$a$ to time period $b$, we state:
\begin{equation}
(c_{ia}=0\land\dots\land c_{ib}=0)\lor(c_{ia}=d_{ia}\land\dots\land
c_{ib}=d_{ib})\footnote{Notice that $d_{ia}=\dots=d_{ib}>0$.}\label{10}
\end{equation}

Finally, the following (obvious) redundant constraints can be added in order to
help orienting the search:

\begin{align}
\forall i\in 1\dots k, j\in 2\dots m:\ &0\leq\bout_{ij}\leq\mathit{TankFlow_i}\label{11}\\
\forall i\in 1\dots k, j\in 2\dots m:\ &\bout_{ij}\leq\buf_{ij-1}\label{12}
\end{align}

Constraints~\ref{1} state that the capacity of the WWTP is not
exceeded at any time. Constraints~\ref{2} and~\ref{3} define the
amount of water inside every buffer at every time interval, taking
into account the amount of water inside each buffer at the previous
time interval, and the current output and input flows for this buffer.
Constraints~\ref{4} require the capacity of each buffer not being
exceeded at any time, and constraints~\ref{5} impose all buffers being
empty at the deadline. Constraints from~\ref{6} to~\ref{9} are
restrictions on the output flow from the buffers (or retention tanks):
the output flow at the first time interval must be zero (as the buffer
is empty) and, at subsequent time intervals, it can be either zero, or
it can be equal to the tank flow (provided that there is enough water
inside the buffer) or it can be equal to the remaining water inside
the buffer if this is less or equal than the tank
flow. Constraints~\ref{10} express the dichotomy of throwing each
discharge to the river or redirecting it to a buffer.

Constraints~\ref{11} and~\ref{12} are unnecessary, but have proved to
be helpful in our experiments. Notice that, although the value of the
$\bout$ variables is perfectly defined by constraints~\ref{6}
to~\ref{9}, restricting the domain of the $\bout$ variables can help
in the search for solutions.

\subsection{IP modeling}\label{IPmodeling}

In order to obtain an IP instance from the previous SMT instance, we need to
convert logical combinations of linear constraints into conjunctions
of linear constraints. We use standard transformations like the ones
of~\cite{Williams1978}.

We define, $\forall i\in 1\dots k, j\in 1\dots m$, zero-one variables
$r_{ij}$ denoting whether discharge from industry $i$ at time period
$j$ is actually scheduled or else redirected to a buffer. Then we
replace $c_{ij}$ with $r_{ij}\cdot d_{ij}$ inside constraints~\ref{1},
~\ref{2} and~\ref{3}. Constraints~\ref{4}, \ref{5} and \ref{6} remain
the same. The zero-one variables $r_{ij}$ allow constraint~\ref{10} to
be translated into
\[
r_{ia}+\dots+r_{ib}=0 \lor r_{ia}+\dots+r_{ib}=b-a+1\text{.}
\]
This can then be encoded as a conjunction of linear constraints by
defining additional zero-one variables $\delta_{iab}$ for every
discharge from an industry $i$ spanning from time period $a$ to time
period $b$, and stating:
\begin{align}
r_{ia}+\dots+r_{ib}+(b-a+1)\cdot\delta_{iab}&\leq b-a+1\\
-(r_{ia}+\dots+r_{ib})-(b-a+1)\cdot\delta_{iab}&\leq -(b-a+1)
\end{align}

The disjunction of constraints~\ref{7}, \ref{8}
and~\ref{9} can be expressed as
\begin{align}
\delta'_{1ij}&\to \bout_{ij}=0\label{15}\\
\delta'_{2ij}&\to \bout_{ij}=\mathit{TankFlow_i} \land
\buf_{ij-1}\geq\mathit{TankFlow_i}\label{16}\\
\delta'_{3ij}&\to \bout_{ij}=\buf_{ij-1} \land
\buf_{ij-1}\leq \mathit{TankFlow_i}\label{17}
\end{align}
where $\delta'_{1ij}$, $\delta'_{2ij}$ and $\delta'_{3ij}$ are again
zero-one variables, and
\begin{equation}
\delta'_{1ij}+\delta'_{2ij}+\delta'_{3ij}\geq 1
\end{equation}
Then constraints~\ref{15}, \ref{16} and~\ref{17} can be transformed
into a conjunction of linear constraints by using $\mathit{Big}-M$
like constraints.\footnote{The idea of $\mathit{Big}-M$ constraints is
  the following: a disjunction like, e.g., $(x\leq 0)\lor b$, where
  $b$ is a propositional variable, can be converted into
  $x\leq\mathit{ubound}(x)b$, where $\mathit{ubound}(x)$ denotes an
  upper bound of $x$.} This way, constraint~\ref{15} becomes
\begin{equation}
\bout_{ij}+\mathit{TankFlow_i}\cdot\delta'_{1ij}\leq\mathit{TankFlow_i}\label{19}
\end{equation}
and constraint~\ref{16} becomes
\begin{align}
\mathit{TankFlow_i}\cdot\delta'_{2ij}-\bout_{ij} &\leq 0\label{20}\\
-\buf_{ij-1}+\mathit{TankFlow_i}\cdot\delta'_{2ij} &\leq 0
\end{align}
Notice that these constraints work in conjunction with
constraints~\ref{11}, which are mandatory here: on the one hand,
from~\ref{19} we get $\bout_{ij}\leq 0$ whenever $\delta'_{1ij}=1$,
which together with $0\leq\bout_{ij}$ (from~\ref{11}) gives us
$\bout_{ij} = 0$ as we need; on the other hand, from~\ref{20} we get
$\mathit{TankFlow_i} \leq \bout_{ij}$ whenever $\delta'_{2ij} = 1$,
which together with $\bout_{ij} \leq \mathit{TankFlow_i}$
(from~\ref{11}) gives us $\bout_{ij} = \mathit{TankFlow_i}$ as we
need.

Finally, constraint~\ref{17} becomes
\begin{align}
\buf_{ij-1}-\bout_{ij}+\mathit{TankCapacity_i}\cdot\delta'_{3ij} &\leq
\mathit{TankCapacity_i}\label{22}\\
\buf_{ij-1}+\mathit{TankCapacity_i}\cdot\delta'_{3ij} &\leq \mathit{TankCapacity_i}+\mathit{TankFlow_i}
\end{align}
Constraints~\ref{12} are mandatory for similar reasons as before,
since they work in conjunction with~\ref{22}.

\section{Benchmarking}\label{benchmarking}

Here we comment on some benchmarking we did, showing that
state-of-the-art SMT solvers outperform best IP solvers with the
previous modeling of the WWTPP.
We worked with two sets of benchmarks, one coming from real data and
another coming from randomly generated data.\footnote{The data used in
  both sets of benchmarks can be found in {\tt
    http://imae.udg.edu/\~{}mbofill/wwtpp.tar}}

In the real set of benchmarks we used data coming from 8 industries
(each one having its own retention tank), with a total of 94
discharges planned within a period of 24 hours. We took a time
discretization of one hour and an overall deadline of 24 hours for the
schedule. Different problem instances were generated with different
capacities of the wastewater treatment plant, ranging from 2000 units
to 10000, at increments of 20. This way, an easy-hard-easy
transition was observed (as already noted
by~\cite{CaseauLaburthe96,HerroelenDeReyck99} for similar scheduling
problems) with a transition from unsatisfiability to satisfiability
taking place at 5000 units of capacity.

For the random set of benchmarks we considered a total of 114
discharges from 10 industries (having again each one an associated
retention tank), all of them being planned within a period of 24
hours. Although randomly generated, both the magnitude and duration of
the discharges and the size of the retention tanks was restricted to
be within reasonable limits. We took a time discretization of one hour
and an overall deadline of 26 hours for the schedule. From this data
different problem instances were generated, with a capacity of the
wastewater treatment plant ranging from 5000 to 30000 units, at
increments of 100, resulting into a transition from unsatisfiability
to satisfiability at 14500 units.

All the benchmarks, written according to the modeling of
Subsection~\ref{SMTmodeling} in the SMT-LIB standard language, were
submitted the last year to the SMT library,\footnote{{\tt
    http://www.smt-lib.org}} and some of them were chosen for the
annual SMT competition\footnote{{\tt http://www.smt-comp.org}} in the
corresponding category.

Table~\ref{naive_ip} shows the percentage of solved benchmarks and the
total time spent by IBM ILOG CPLEX 11, Z3.2$^\alpha$ (SMT-COMP'08
QF\_LIA division winner) and Yices 2 (SMT-COMP'09 QF\_LIA division
winner), with a time out of 1800 seconds for each instance in the real
set, and of 300 seconds in the random set.  All benchmarks were
executed on a 3.80~GHz Intel Xeon machine with 3.5 GB of RAM running
under GNU/Linux 2.6. The modeling given in Subsection~\ref{IPmodeling}
was used for CPLEX.

\ctable[
caption = SMT vs. IP,
label = naive_ip,
]{ccccc}{
\tnote[a]{Minimizing sum of buffer contents.}
\tnote[b]{Without objective function.}
}{
\FL
& \multicolumn{2}{c}{Real set} & \multicolumn{2}{c}{Random set} \NN
Solver & \% Solved & Time & \% Solved & Time \ML
Yices& 100.0 & 1227.4 & 100.0 & 5.2  \NN
Z3& 99.8 & 1152.7 & 100.0 & 285.2 \NN
CPLEX\tmark[a] & 97.6 & 1855.5 & 98.8  & 594.8 \NN
CPLEX\tmark[b] & 93.5 & 1811.5 & 92.9 & 25.0  \LL
}

As it can be seen, state-of-the-art SMT solvers clearly outperform
CPLEX on this benchmarks. It is specially remarkable that Yices solves
all the benchmarks, and Z3 only fails in solving one from the real set
around the phase transition. Moreover, Yices is able to solve all the
251 benchmarks from the random set in only 5.15 seconds, being almost
insensitive to the phase transition. W.r.t.\ CPLEX, although it has
very good performance in many instances, it fails to solve some of
them around the phase transition.  Since SMT solvers, like it does
CPLEX, use a simplex procedure for handling atomic linear constraints,
other elements of SMT technology such as conflict-driven lemma
learning, backjumping or restarts can be playing a central role in
this problem. 

It is worth noting that worse results are obtained by CPLEX if no
objective function is used. After trying with several objective
functions, we obtained the best results by minimizing the sum of
buffer contents. This somehow corresponds to an eager strategy
consisting in avoiding the use of buffers if possible (and hence
prioritizing discharges of wastewater at their preliminarily scheduled
times) and emptying the buffers as soon as possible. Notice however
that no objective function or user-given search strategy is possible
with SMT solvers, which are completely black-box for the user and,
still, better results are obtained.

\section{Comparison with Constraint Programming}\label{cp}

For the sake of completeness, in this section we detail the results
obtained with several Constraint Programming (CP) tools on our benchmarks.

In order to do the benchmarking, our modeling needs to be translated
into several CP dialects. For the comparison to be fair, in all cases
we must choose an encoding as similar as possible to the one described
in Subsection~\ref{SMTmodeling}. This implies avoiding the use of
global constraints and sophisticated search strategies that can be
available in CP tools. For this reason, we have only used labeling
strategies.\footnote{Notice that there is always a default labeling
  strategy in these tools and, hence, trying with some labeling
  options does not imply doing any change in the encoding.}  Results
on a different encoding, using the {\tt cumulative} global constraint,
are given in the next section.

Since the translation of the encoding described in
Subsection~\ref{SMTmodeling} into a CP program over finite domains is
almost direct, the encodings obtained for each CP tool are very
similar and hence we do not detail them here. Moreover, for
solvers providing a FlatZinc front-end, we have used the same MiniZinc
model: MiniZinc~\cite{MiniZinc07} proposes to be a standard CP
modeling language that can be translated into an intermediate language
called FlatZinc. FlatZinc instances can be obtained from MiniZinc
instances by using the MiniZinc-to-FlatZinc translator {\tt mzn2fzn},
and then can be plugged into any solver providing an specialized
front-end for FlatZinc.

\ctable[
caption = SMT vs. CP,
label = naive_cp,
]{lcccc}{
\tnote[a]{With labeling options: max, down.}
\tnote[b]{Using CP engine.}
\tnote[c]{Using LP engine.}
}{
\FL
& \multicolumn{2}{c}{Real set} & \multicolumn{2}{c}{Random set} \NN
Solver & \% Solved & Time & \% Solved & Time \ML
SICStus\tmark[a] & 68.8 & 258.9 & 81.7 & 27.7 \NN
Comet\tmark[b] & 76.3 & 744.5 & 53.8 & 196.0 \NN
Comet\tmark[c] & 46.4 & 43.8 & 71.7 & 27.1 \NN
Tailor $+$ Minion & 81.3 & 547.6 & 44.6 & 98.3 \NN
mzn2fzn $+$ G12 & 28.9 & 32.2 & 74.9 & 77.1 \NN
mzn2fzn $+$ Gecode & 0.0 & 0.0 & 37.1 & 9.8 \NN
mzn2fzn $+$ ECL$^i$PS$^e$& 0.0 & 0.0 & 0.0 & 0.0 \NN
mzn2fzn $+$ SICStus & 0.0 & 0.0 & 37.1 & 345.8 \NN
mzn2fzn $+$ fzn2smt $+$ Z3 & 99.8 & 4735.8 & 100.0 & 159.0 \NN
mzn2fzn $+$ fzn2smt $+$ Yices & 99.8 & 702.8 & 100.0 & 40.5 \LL
}
 
Table~\ref{naive_cp} shows the results obtained by several CP solvers
on the benchmarks described in Section~\ref{benchmarking}, except for
the last two entries, which show the results obtained by the same SMT
solvers used in Section~\ref{benchmarking}, but where SMT instances
have been obtained from FlatZinc instances through an experimental
compiler {\tt fzn2smt}.\footnote{Available at {\tt
    http://imae.udg.edu/recerca/lap}} The table refers only
to the solving time (we do not include translation times since we are
interested in comparing solving times, regardless of the input
language). All benchmarks were executed on a 3~GHz Intel Core 2 Duo
machine with 1 GB of RAM running under GNU/Linux 2.6.

At a first glance we can observe that SMT solvers are far better than
other tools on these benchmarks. It is remarkable that, after the two
step translation from MiniZinc-to-FlatZinc-to-SMT, we obtain similar
(and in some case even better) results to the ones in
Section~\ref{benchmarking}.

We tried different labeling strategies with CP solvers, but almost
identical results were obtained. Hence, unless contrarily indicated,
the results in Table~\ref{naive_cp} are for the default strategy,
which is usually first-fail: selecting the leftmost variable with
smallest domain next, in order to detect infeasibility early. This is
often a good strategy. However, with SICStus Prolog we obtained
significantly better results when using the {\tt max} and {\tt down}
options: selecting the leftmost variable with the greatest upper bound
next, and exploring its domain in descending order. In our program,
this translates to a strategy consisting in giving priority to the
biggest discharges, and keeping them in buffers as least as
possible. Notice that this roughly coincides with the objective
function giving best results in the IP approach of
Section~\ref{benchmarking}.

The concrete versions of the CP solvers we used are: SICStus Prolog 4.0.1
(for the first entry in the table), SICStus Prolog 4.1.1 (with
FlatZinc support, for the MiniZinc case), Comet 2.0, Minion 0.9, G12
MiniZinc 1.0.3, Gecode 3.2.2, and ECL$^i$PS$^e$ 6.0.  For the case of
Minion, we used Tailor as a translator from the
\essence~\cite{Essence08} high-level language to the Minion language,
in the same spirit of using the Minizinc-to-Flatzinc translator {\tt
  mzn2fzn}. This allowed us to use an almost identical model. Comet
already supports a high-level language which allowed us to express the
constraints in a very similar way. Moreover, for the case of Comet we
tried both the CP engine and the LP engine, with no clear winner.
We want to remark that we are aware of IBM ILOG CP Optimizer, which
uses constraint programming to solve detailed scheduling problems and
combinatorial problems not easily solved using mathematical
programming methods. Unfortunately we were not able to test this tool
on our benchmarks, since the trial version has severe limitations in
the number of variables and in the number of allowed constraints.

\section{A different approach for Constraint Programming}\label{cum}

An alternative approach is to solve the WWTPP by exploiting the use of
the {\tt cumulative} constraint within a CP system, since this
constraint is closely related to our problem.  Many CP systems, such
as CHIP V5, ECL$^i$PS$^e$, B-Prolog and SICStus Prolog, include the
\texttt{cumulative} global constraint in their finite domain
library. This constraint was originally introduced into the CHIP
programming system to describe and solve complex scheduling problems
\cite{AggounBeldiceanu93a}. Its habitual syntax is
\texttt{cumulative(Starts,Du\-rations,Resources,Limit)}, where
\texttt{Starts}, \texttt{Durations}, and \texttt{Resources} are
lists of integer domain variables or integers of the same length, and
\texttt{Limit} is an integer. The declarative meaning is: if the lists
denote respectively the start times, durations and resource capacity
requirements of a set of tasks, then the sum of resource usage of all
the tasks does not exceed \texttt{Limit} at any time. One should
expect that, by using this constraint adequately, the performance of a
CP system on the previous problem will be better (or, at least, not
worse) than if not using it.

Our modeling using the {\tt cumulative} constraint goes as follows.
Given a discharge $i$ of duration $d_i$ and resource capacity
requirement $c_i$, since it can either go directly to the river or be
redirected to a retention tank of certain output rate $r$, we create a
set of new $n$ discharges of duration $1$ and capacity requirement
$r$, and one discharge of duration $1$ and non-negative requirement
capacity $r'\leq r$ (the remainder), such that $d_i c_i = r n +
r'$. Observe that by dividing the discharges into a number of
discharges of duration $1$ we get rid of preemption.  Then, by using
reified constraints, we state that the capacity requirements of those
$n+1$ new discharges is actually $0$ if and only if the associated
original discharge $i$ goes to the river.

Notice that a set of remainders (each of them coming from a different
original discharge of the same industry) could eventually be
redistributed, forming a new set of discharges of resource capacity
$r$ plus one single remainder. However, such redistribution should be
made for the remainders being available at each time, i.e.,
dynamically, and this does not go in the direction of an encoding
using the {\tt cumulative} constraint, which requires a fixed set of
resources. Therefore, here we do not consider the possibility of
redistributing the remainders. Although this is a inexact formulation
of the problem, in practice it results a very few times in a smaller
set of solutions than with the encoding used in the previous
sections. And, in any case, since this simplification results in a
smaller search space, it is likely to favour this approach.

Then, apart from stating the obvious release time, precedence and
finishing time constraints, we use the {\tt cumulative} constraint
two-fold.  On the one hand, we use it in order to assure that the WWTP
capacity is not exceeded. On the other hand, we use it in order to
assure that the output rate and capacity of every retention tank is
not exceeded. This second use implies stating two {\tt cumulative}
constraints for each industry, in the following way:

Let \texttt{[I1,\dots,In]} be a list with the initial times of the
discharges kept in the retention tank of a industry, let
\texttt{[H1,\dots,Hn]} be the times at which they are respectively
flushed out from the tank, let \texttt{[C1,\dots,Cn]} be their
resource capacity requirements, let \texttt{r} be the output rate of
the tank, and let \texttt{c} be the capacity of the tank. Then we
state
\begin{center}
{\tt cumulative([H1,\dots,Hn],[1,\dots,1],[C1,\dots,Cn],r)}
\end{center}
in order that the output rate of the tank is not
exceeded, and 
\begin{center}
{\tt cumulative([I1,\dots, In],[H1-I1,\dots,Hn-In],[C1,\dots,Cn],c)}
\end{center}
in order that the capacity of the tank is not exceeded.

Finally, for symmetry breaking, we state ordering constraints between
(indistinguishable) discharges from each retention tank. Since all
these discharges are of duration $1$, this improvement dramatically
reduces the search space.

\ctable[
caption = Cumulative modeling,
label = cumulative,
]{lcccc}{
\tnote[a]{With labeling options: max, down.}
}{
\FL
& \multicolumn{2}{c}{Real set} & \multicolumn{2}{c}{Random set} \NN
Solver & \% Solved & Time & \% Solved & Time \ML
SICStus\tmark[a] & 76.6 & 3231.9 & 95.2 & 1347.2 \NN
mzn2fzn $+$ G12 & 67.6 & 64.0 & 12.8 & 8.5 \NN
mzn2fzn $+$ Gecode & 72.6 & 1709.2 & 61.4 & 24.8 \NN
mzn2fzn $+$ ECL$^i$PS$^e$ & 23.2 & 3255.7 & 17.9 & 637.0 \NN
mzn2fzn $+$ SICStus & 69.3 & 2029.8 & 51.4 & 432.4 \LL
}

Table~\ref{cumulative} shows the results obtained by the CP solvers
supporting the {\tt cumulative} global constraint on the same
benchmarks as in the previous sections. Again, we used the possibility
of sharing a unique MiniZinc model, except for the first entry,
where we directly built a Prolog program.  We can observe that, in
general, the results are better than with the previous encoding for
the same CP solvers (with the only exception of G12 in the random
set). However, these results are still far from the ones obtained by
SMT solvers. This can be due to the fact that we are using two {\tt
  cumulative} constraints for each industry (for assuring,
respectively, that the output rate and the capacity of each retention
tank is not exceeded), plus one {\tt cumulative} global constraint
(for assuring that the WWTP capacity is not exceeded) and, moreover,
we are using many reified constraints (for the dichotomy of sending
the discharges either to the river or to a retention tank), making
thus difficult for the CP solvers to take profit of their algorithms
for the {\tt cumulative} constraint.

\section{Conclusion and future work}\label{conclusion}

In this paper we have presented the Wastewater Treatment Plant Problem
(WWTPP), a real problem in the cumulative scheduling class, and have
compared several techniques for solving it. The encoding of the WWTPP
into SAT modulo linear integer arithmetic, and using a
high-performance SMT solver as a black-box for solving it, has
turned out to be one of the best approaches. Specifically, we have
seen that state-of-the-art SMT solvers are competitive with current
best IP solvers, en even better on difficult instances of this problem
(i.e., the ones around the phase transition). We think that this is a new
result in the direction of showing that current SMT solvers are ready
to solve real problems outside the verification area, and that they
provide a nice compromise between expressivity and efficiency for
scheduling problems.

Let us recall that SMT solvers, like IP tools, use a simplex procedure
for handling atomic linear constraints. However, the particular
treatment of bound constraints of the form $x\leq k$ or $x\geq k$
inside a simplex procedure like the one of Yices, must be a key
ingredient for the good results obtained in this problem by this
solver (notice that many constraints in this problem are of this
form). Also, we think that usual SMT techniques such as backjumping,
restarts, and conflict-driven lemma learning must be a key ingredient
for the good results obtained around the phase transition.

Moreover, in our point of view, the encoding of the WWTPP as an SMT
problem is simpler than as an IP problem (where logical combinations
of linear constraints must be translated into conjunctions of linear
constraints, with the addition of zero-one variables). Compared to CP,
the SMT approach is not that simple (since most CP tools provide a
high-level language front-end), but far more efficient. The
performance of SMT solvers on this problem is still more significant
if we take into account that they are completely black-box, and one
cannot provide neither labeling strategies nor local search algorithms
for guiding the search.

It was not our aim to find specialized algorithms for this problem,
but to solve it from a non-expert perspective. In this sense, from our
result it follows that compilers translating from a high-level
language to SMT would be desirable, especially for people having no
experience with constraint satisfaction problems.  Let us recall,
that, after a two step translation from MiniZinc-to-FlatZinc-to-SMT,
we obtained similar (and in some case even better) results to the ones
with a direct SMT encoding.\footnote{Available as new benchmarks at the
  SMT-LIB ({\tt http://www.smt-lib.org}) in the QF\_LIA
  category.} Hence, we think that user-friendly and at the same time
competitive CP tools could arise from such or similar combinations.
In fact, some research has already been done in this direction. For
instance, in~\cite{Bofilletal2009ModRef}, excellent results have been
obtained on benchmarks from the CSPLib~\cite{csplib}, after modeling
them in a high-level language and automatically translating them into
SAT modulo unquantified linear integer arithmetic. The {\tt
  fzn2smt}\footnote{Available at {\tt
    http://imae.udg.edu/recerca/lap}} compiler is a new tool
in the same direction.

On the other hand, as pointed out in~\cite{Nieuwenhuisetal2007RTA},
for dense difference logic problems such as the ones coming from
scheduling, there is still room for improvement in the context of
SMT. Although such an improvement could be done in the solvers for the
usual SMT theories, an alternative interesting thing to do would be to
implement, e.g., a solver for the theory of {\tt cumulative}, to be
used with a modular SMT solver like the ones based on the DPLL(X)
approach~\cite{NieuwenhuisetalJACM2006}.  Notice that this combination
would give us a framework similar to the one of constraint
programming, but with backjumping and learning.
In fact, excellent results have been recently
obtained in~\cite{SchuttFSW09} by modeling the {\tt cumulative}
constraint by descomposition, and by exploiting the autonomous search
and learning capabilities of a SAT solver in a way which resembles
what is done in SMT. This makes us expect good performance results for
SMT solvers in constraint satisfaction problems if backtrackable and
incremental solvers for theories like {\tt cumulative}, {\tt
  alldifferent}, etc, are developed. In this sense, we totally agree
with challenge~12 in~\cite{Nieuwenhuisetal2007RTA}: to develop an SMT
system with the advantages of one of CP's sophisticated global
constraint propagation algorithms and the robustness and efficiency of
SAT's backjumping, lemmas and heuristics, by expressing the global
constraints as a theory.



\bibliographystyle{alpha}

\newcommand{\etalchar}[1]{$^{#1}$}

\end{document}